\documentclass[journal]{IEEEtran}
\makeatletter\if@twocolumn\PassOptionsToPackage{switch}{lineno}\else\fi\makeatother

\usepackage{graphicx}
\usepackage[T1]{fontenc}
\usepackage{wrapfig}
\usepackage{enumerate}

\usepackage{bbding}
\usepackage{multirow}
\usepackage{amsmath}
\usepackage{caption}
\usepackage{subcaption}
\usepackage{tikz}
\usepackage{amssymb}
\usepackage{pifont}
\usepackage{mathtools}
\usepackage{algorithm}
\usepackage{algpseudocode}
\usepackage{cases}
\usepackage{graphics}
\usepackage[outline]{contour}
\usepackage{url,morefloats,floatflt,cancel,tfrupee}
\usepackage[flushleft]{threeparttable}
\usepackage{stfloats}
\usepackage[export]{adjustbox}
\newcommand{\cmark}{\ding{51}}%
\newcommand{\xmark}{\ding{55}}%

\usepackage{color}
\usepackage{pgfplots}
\pgfplotsset{compat=1.13}
\usetikzlibrary{arrows.meta}
\usetikzlibrary{patterns}

\usepackage{scalerel}
\usetikzlibrary{svg.path}
\usepackage[switch]{lineno}
\usepackage{float}
\usepackage{tabularray}

\definecolor{orcidlogocol}{HTML}{A6CE39}
\tikzset{
  orcidlogo/.pic={
    \fill[orcidlogocol] svg{M256,128c0,70.7-57.3,128-128,128C57.3,256,0,198.7,0,128C0,57.3,57.3,0,128,0C198.7,0,256,57.3,256,128z};
    \fill[white] svg{M86.3,186.2H70.9V79.1h15.4v48.4V186.2z}
                 svg{M108.9,79.1h41.6c39.6,0,57,28.3,57,53.6c0,27.5-21.5,53.6-56.8,53.6h-41.8V79.1z M124.3,172.4h24.5c34.9,0,42.9-26.5,42.9-39.7c0-21.5-13.7-39.7-43.7-39.7h-23.7V172.4z}
                 svg{M88.7,56.8c0,5.5-4.5,10.1-10.1,10.1c-5.6,0-10.1-4.6-10.1-10.1c0-5.6,4.5-10.1,10.1-10.1C84.2,46.7,88.7,51.3,88.7,56.8z};
  }
}
\newcommand\orcidicon[1]{\href{https://orcid.org/#1}{\mbox{\scalerel*{
\begin{tikzpicture}[yscale=-1,transform shape]
\pic{orcidlogo};
\end{tikzpicture}
}{|}}}}

\newif\ifpreprint

\ifpreprint
\definecolor{revised}{rgb}{0,0,0} 
\else
\definecolor{revised}{rgb}{0.1,0.1,0.6} 
\fi

\definecolor{huai}{rgb}{0.1,0.1,0.6}
\definecolor{hai}{rgb}{0.1,0.1,0.6}
\definecolor{hung}{rgb}{0.06,0.2,0.65}
\definecolor{ty}{rgb}{0.1,0.1,0.6}

\ifCLASSINFOpdf

\else

\fi

\usepackage{amsmath}
\usepackage{algorithm}
\usepackage{algpseudocode}

\usepackage{mathtools}

\usepackage{url,multirow,morefloats,floatflt,cancel,tfrupee}

\ifpreprint
\else
\fi

\usepackage{adjustbox}
\usepackage{hyperref} 
\usepackage{kbordermatrix} 

\makeatletter

\AtBeginDocument{\@ifpackageloaded{textcomp}{}{\usepackage{textcomp}}}
\makeatother
\usepackage{colortbl}
\usepackage{xcolor}

\makeatletter

\def\mcWidth#1{\csname TY@F#1\endcsname+\tabcolsep}

\def\cAlignHack{\rightskip\@flushglue\leftskip\@flushglue\parindent\z@\parfillskip\z@skip}
\def\rAlignHack{\rightskip\z@skip\leftskip\@flushglue \parindent\z@\parfillskip\z@skip}

\@ifundefined{etal}{\def\etal{\textit{et~al}}}{}

\usepackage{ifxetex}
\ifxetex\else\if@twocolumn\@ifpackageloaded{stfloats}{}{\usepackage{dblfloatfix}}\fi\fi

\AtBeginDocument{
\expandafter\ifx\csname eqalign\endcsname\relax
\def\eqalign#1{\null\vcenter{\def\\{\cr}\openup\jot\m@th
  \ialign{\strut$\displaystyle{##}$\hfil&$\displaystyle{{}##}$\hfil
      \crcr#1\crcr}}\,}
\fi
}

\AtBeginDocument{%
  \@ifpackageloaded{endfloat}%
   {\renewcommand\efloat@iwrite[1]{\immediate\expandafter\protected@write\csname efloat@post#1\endcsname{}}}{\newif\ifefloat@tables}%
}%

\def\BreakURLText#1{\@tfor\brk@tempa:=#1\do{\brk@tempa\hskip0pt}}
\let\lt=<
\let\gt=>
\def\processVert{\ifmmode|\else\textbar\fi}

\@ifundefined{subparagraph}{
\def\subparagraph{\@startsection{paragraph}{5}{2\parindent}{0ex plus 0.1ex minus 0.1ex}%
{0ex}{\normalfont\small\itshape}}%
}{}

\newcommand\role[1]{\unskip}
\newcommand\aucollab[1]{\unskip}
  
\@ifundefined{tsGraphicsScaleX}{\gdef\tsGraphicsScaleX{1}}{}
\@ifundefined{tsGraphicsScaleY}{\gdef\tsGraphicsScaleY{.9}}{}
\def\checkGraphicsWidth{\ifdim\Gin@nat@width>\linewidth
	\tsGraphicsScaleX\linewidth\else\Gin@nat@width\fi}

\def\checkGraphicsHeight{\ifdim\Gin@nat@height>.9\textheight
	\tsGraphicsScaleY\textheight\else\Gin@nat@height\fi}

\def\fixFloatSize#1{}
\let\ts@includegraphics\includegraphics

\def\inlinegraphic[#1]#2{{\edef\@tempa{#1}\edef\baseline@shift{\ifx\@tempa\@empty0\else#1\fi}\edef\tempZ{\the\numexpr(\numexpr(\baseline@shift*\f@size/100))}\protect\raisebox{\tempZ pt}{\ts@includegraphics{#2}}}}

\AtBeginDocument{\def\includegraphics{\@ifnextchar[{\ts@includegraphics}{\ts@includegraphics[width=\checkGraphicsWidth,height=\checkGraphicsHeight,keepaspectratio]}}}

\DeclareMathAlphabet{\mathpzc}{OT1}{pzc}{m}{it}

\def\URL#1#2{\@ifundefined{href}{#2}{\href{#1}{#2}}}

\def\UrlOrds{\do\*\do\-\do\~\do\'\do\"\do\-}%
\g@addto@macro{\UrlBreaks}{\UrlOrds}

\edef\fntEncoding{\f@encoding}

\makeatother

\newif\ifmultipleabstract\multipleabstractfalse%
\makeatletter
\AtBeginDocument{\@ifpackageloaded{longtable}{%
\def\LT@makecaption#1#2#3{%
  \LT@mcol\LT@cols c{\hbox to\z@{\hss\parbox[t]\LTcapwidth{%
    \sbox\@tempboxa{#1{#2: } #3}%
    \ifdim\wd\@tempboxa>\hsize
      #1{#2: }\textsc{#3}%
    \else
      \hbox to\hsize{\hfil\box\@tempboxa\hfil}%
    \fi
    \endgraf\vskip\baselineskip}%
  \hss}}}
}{}}
\makeatother

\usepackage{xspace}
\makeatletter
\DeclareRobustCommand\onedot{\futurelet\@let@token\@onedot}
\def\@onedot{\ifx\@let@token.\else.\null\fi\xspace}

\def\ie{\emph{i.e}\onedot}

\def\etal{\emph{et al}\onedot}
\makeatother

\begin{document}
\include{textbox} 
\title{A novel framework for adaptive stress testing of autonomous vehicles in multi-lane roads}


\author{{Linh~Trinh}, {Quang-Hung~Luu}, {Thai~M.~Nguyen} and {Hai~L.~Vu\textsuperscript{\Envelope}}
\thanks{Corresponding authors: Prof. Hai L. Vu (hai.vu@monash.edu).}
}

\ifpreprint
\markboth{Preprint}{}
\else
\markboth{JOURNAL OF LATEX CLASS FILES, VOL. XX, NO. X, XX 202X}{}
\fi

\maketitle 

\IEEEpeerreviewmaketitle

\begin{abstract}
Stress testing is an approach for evaluating the reliability of systems under extreme conditions which help reveal vulnerable scenarios that standard testing may overlook. Identifying such scenarios is of great importance in autonomous vehicles (AV) and other safety-critical systems. Since failure events are rare, naive random search approaches require a large number of vehicle operation hours to identify potential system failures. Adaptive Stress Testing (AST) is a method addressing this constraint by effectively exploring the failure trajectories of AV using a Markov decision process and employs reinforcement learning techniques to identify driving scenarios with high probability of failures. However, existing AST frameworks are able to handle only simple scenarios, such as one vehicle moving longitudinally on a single lane road which is not realistic and has a limited applicability. In this paper, we propose a novel AST framework to systematically explore corner cases of intelligent driving models that can result in safety concerns involving both longitudinal and lateral vehicle's movements. Specially, we develop a new reward function for Deep Reinforcement Learning to guide the AST in identifying crash scenarios based on the collision probability estimate between the AV under test (\ie, the ego vehicle) and the trajectory of other vehicles on the multi-lane roads. To demonstrate the effectiveness of our framework, we tested it with a complex driving model vehicle that can be controlled in both longitudinal and lateral directions. Quantitative and qualitative analyses of our experimental results demonstrate that our framework outperforms the state-of-the-art AST scheme in identifying corner cases with complex driving maneuvers.

\end{abstract}

\section{Introduction}\label{section:intro}
Stress testing is a commonly used method being developed to assess the performance and resilience of systems under extreme conditions \cite{stress_test_1,stress_test_2,stress_test_3,stress_test_4}. Unlike normal testing, which explore the system's functionality under expected operational scenarios, stress testing pushes systems beyond their typical limits to identify potential failure to understand their behavior under adverse circumstances. This approach is essential for ensuring reliability, and robustness of safety-critical systems. The main challenge is that the analysis of such systems, like aircraft collision avoidance systems \cite{ast_AirCraff2015}, autonomous vehicles \cite{ast4av2018} and autonomous shipping \cite{ast4usv}, that operate over a sequence of time steps requires finding the most likely path to a set of failure states. The complex stochastic environment in which the system operates makes it impossible to completely eliminate failures in many applications, like autonomous driving. Therefore, safety validation seeks to determine not only if a failure is possible, but also which failures are likely to occur. 

Recently, an emerging approach being referred to as adaptive stress testing (AST) was introduced from the bottom up perspective \cite{ast4av_rewardAug2019,ast_ma@AAAI2023,ast_ma2022}. 
The goal of this approach is to identify potential failure scenarios and their likelihoods of occurrences. In particular, it is designed to discover the most probable system failure events, which are traditionally formulated as a Markov Decision Process (MDP) \cite{ast_AirCraff2015,ast4av2018}. In this process, Reinforcement learning (RL) is often employed in conjunction with domain knowledge to efficiently explore the vast state space and identify failure events. To this end, the search process is guided by a reward function that encourages failures such as collisions and near-misses with a high transition probability.

Autonomous vehicles (AVs) are expected to serve as an important component of modern transportation systems. Underpinned by the intelligent capability, the advancement of AVs has been driven by non-traditional automotive companies including Tesla, Alphabet, Waymo, among others \cite{validation_odd,scenario_pdf,euro_ncap_2023}. The intelligence of AV is classified into six different levels, from 0 (zero automation) to 5 (full self-driving) \cite{odd@sae} where each level is equipped with a certain degree of Advanced Driver-Assistance System (ADAS) or Automated Driving System (ADS) features such as Adaptive Cruise Control (ACC), Lane Keeping Assist (LKA), Highway Pilot (HWP), and others \cite{nhtsa_fw,odd@sae}. At the same time, standards and regulations such as ISO 26262, 21448, ANSI/UL 4600 \cite{iso2018_26262,iso2020tr,iso21448,ansi_ul_4600}, and national regulators such as the USA NHTSA, EURO NCAP \cite{nhtsa_fw,euro_ncap_2018}, ASEAN NCAP \cite{asean_ncap}, have been revised to guarantee the safety of AVs.

That is to say ensuring the safe operations of AVs, a typical cyber-physical system (CPS), is essential for their broad acceptance and public trust. In particular, verifications and validation are crucial to the safety of AV systems at all stages of their development from simulations to on-road testing. While on-road testing is eventually required for all AVs, it is time-consuming and costly, especially considering the wide range of scenarios that they need to cover. Such challenges make it difficult to thoroughly validate all scenarios that an AV may encounter \cite{intro_driving_safety}. In contrast, a benefit of simulation-based testing is that it is generally much cheaper to conduct while there is no safety concern to the public. \cite{intro_challenge_test}. Having said that, existing simulation-based tests only have a limited set of scenarios using naive random search approaches which may fail to provide adequate coverage of the entire space due to the scarcity of dangerous scenarios.

To address this challenge, in the literature, AST framework has recently been used to explore the failure patterns of the intelligent driving model (IDM) under test \cite{review_2023}. Prior to the application of self-driving in traffic, AST has shown great successes in numerous other relating applications, including the evaluation of aircraft collision avoidance \cite{ast_AirCraff2015}, maritime collision avoidance \cite{ast_ma2022,ast_ma@AAAI2023}, and autonomous vehicle pedestrian collision avoidance \cite{ast4av2018,ast_formulation2020, ast4av_rewardAug2019, ast4av_simulation2019,ast4av_fidelity2021}.

Nevertheless, the existing AST-based studies for AV driving have several limitations. First, the scenarios used for testing are basic, which does not include complexity of real-world driving situations. For example, \cite{ast4av2018} used a simple scenario, in which the intelligent ego vehicle approaching a crosswalk where a pedestrian is crossing in front of it, to test an IDM that monitors the closest traffic participant to determine a safe distance and imminent collisions without having or taking into account the actions of surrounding vehicles. 
Second, the traditional IDMs under test are simple and do not account for sophisticated, realistic behaviours in driving. Furthermore, from level 1 to level 5 of self-driving, the majority of ADAS/ADS features, such as LKA, HWP, ACC, are based on the capability to assist the AV operating on multi-lane roads. They consist of highways, multi-lane urban roads, intersections, curves, bridge roads, and others. Therefore, developing a test method that can be applied to multi-lane road environments involving both longitudinal and lateral movements is of great importance from both theory and application perspectives.

In this study, we propose a novel AST framework to address the above-mentioned limitations. Our main contributions are:
\begin{itemize}
\item Propose a novel AST framework to stress test the AV in a complex multi-lane road environment which includes a new reward function which encourages safe driving among other vehicles in the multi-lane road while supporting the identification of potential crashed scenarios.
\item Develop a unified intelligent driving model (uIDM) that facilitates the movement of AV in both longitudinal and lateral directions and set it to be the target system under test which is more realistic than models used in the existing literature.
\item Calibrate the framework using observations from California's accident reports and then assess its performance against the common existing IDM model in order to gain insights into the effectiveness and the efficiency of the proposed framework.
\end{itemize}

The rest of our paper is organized as follows. 
In Section~\ref{section:related_works}, we review related studies of AST and intelligent vehicles, discuss the existing approaches and highlight research gaps in this area together with a background of the method used. In Section~\ref{section:method}, we describe our framework including the development a comprehensive intelligent driving model and novel reward function to identify potential corner cases using the proposed intelligent driving model in complex multi-lane road traffic. Experiment setup and calibration of reward function is described in Section~\ref{section:exp}. In Section~\ref{section:result}, we describe the main results with quantitative and qualitative analyses to demonstrate the effectiveness of our framework and further discuss their insights. Finally, we conclude the main findings of our paper in Section~\ref{section:conclusion}.


\section{Related works and Preliminaries}\label{section:related_works}
\subsection{Related works}
\begin{table*}[tbhp]
\centering
\caption{Comparison of related works using Adaptive Stress Testing for autonomous vehicle (Long.: longitudinal movement; Lat.: lateral movement; GT: ground truth; SUT: system under test; '-': not considered.).}\label{table:compare_related}

\begin{center}
\begin{tabular}{|l|c|c|c|c|c|l|c|} \hline
\multirow{2}{*}{\textbf{Works}} & \multicolumn{2}{c|}{\textbf{SUT}} & \multicolumn{5}{c|}{\textbf{Adaptive Stress Test framework}}  \\ 
\cline{2-8} 
& Long.& Lat. & Collision & Traffic objects                            & Lane & Major factors of reward   & Not rely on GT \\
\hline\hline 
\begin{tabular}[t]{@{}l@{}}
\cite{ast4av2018},\cite{ast_formulation2020}, \cite{ast4av_ethic2022}, \cite{ast4av_fidelity2021}, \cite{ast_ma2022}, \cite{ast_ma@AAAI2023}
\end{tabular}              & {\cmark } & {\xmark }    & Front         & 1-pedestrian                                                       & None             & \begin{tabular}[t]{@{}l@{}}- Path likelihood\\ - Failure event\end{tabular}                                & {\cmark }                 \\  \hline

\cite{ast4av_rewardAug2019}              & {\cmark } & {\xmark }  & Front         & $n$-pedestrians                                                       & None             & \begin{tabular}[t]{@{}l@{}}- Responsibility-sensitive safety\\ - Trajectory dissimilarity\end{tabular}                                 &  {\cmark }                  \\  \hline

\cite{ast4av_laneChange2021}             & - & - & -      & $n$-vehicles                                                       & Multiple             & 
\begin{tabular}[t]{@{}l@{}}- Failure event\\  - Human based Critical State Classifier\end{tabular}                                 &  {\xmark }                  \\  \hline\hline 

{Our framework}             & {\cmark } & {\cmark }  & \begin{tabular}[t]{@{}l@{}}- Front\\  - Side \\ - Rear 
safety\end{tabular}      & $n$-vehicles & Multiple             & \begin{tabular}[t]{@{}l@{}}- Failure event\\  - Ego vehicle 
collision\\ - Other vehicle 
safety
\end{tabular} & {\cmark }       \\  \hline          
\end{tabular}
\end{center}
\end{table*}

There have been numerous efforts and studies aimed at developing intelligent or realistic driving systems for autonomous or self-driving vehicles, \ie, AVs. Several studies \cite{rel_dynamic_conf,review_drl4av_2021,review_dl4av_2020,review_decision_planing4av_2018,review_motion_planning4av_2023}, have focused on developing intelligent path planning, motion control or decision making techniques with promising results. One major remaining challenge is to ensure safety while operating in autonomous mode. AV manufacturers often use public road testing to detect and improve AV driving safety performance \cite{rl_test_scenario} within its targeted Operational Design Domain (ODD) \cite{odd@sae}, but road trialling can be hazardous costly and time-consuming \cite{review_test_av}. Identifying scenarios (or corner cases) using simulation where self driving might cause safety concerns has emerged as a good alternative \cite{review_blacbox_safe_test}, which is safer and more affordable than the field-testing. Furthermore, simulation enables the exploration of critical scenarios in large number and at scale, which are difficult to conduct in real-world road trials.

Adaptive stress testing has attracted noticeable successes in understanding and advancing intelligent systems \cite{ast4av_simulation2019,ast4av_rewardAug2019,ast4av_perception2022,ast_ma@AAAI2023,ast_AirCraff2015,ast_ma2022}. It has been used to test various collision avoidance systems for aircraft, maritime and transportation applications \cite{ast_AirCraff2015,ast_ma2022}. One of the early application of AST is for aircraft collision avoidance systems \cite{ast_AirCraff2015}. It helped identify potential failure events by simulating diverse sequences of disturbances leading to near mid-air collisions using the Monte Carlo Tree Search (MCTS) methodology \cite{mcts}. This search strategy was directed towards sequences of failures that were similar to the most promising ones identified so far, while also exploring the less likely ones. Another application domain of AST is maritime collision avoidance \cite{ast_ma2022} where it was used to uncover failures associated with collisions with adversary vessels. The approach that leverages MCTS is constrained by significant computations required for searching and solving. Lytskjold and Mengshoel \cite{ast_ma@AAAI2023} attempted to address this limitation by reducing the search time of MCTS with neural networks, and demonstrated its improvement performance in their recent work.

In transport research, AST has been applied to reveal the failures of Intelligent Driving Model (IDM) used in modeling self-driving vehicles. One early work \cite{ast4av2018} tested the ego vehicle against the scenario with a pedestrian crossing the road. The AV under test underpinned by an IDM which maintained a safe distance from the nearest vehicle or pedestrian to avoid any collisions. Two different families of AST methods were adopted, namely, Monte Carlo Tree Search (MCTS) and Deep Reinforcement Learning (DRL), allowing them to compare the effectiveness of different machine learning algorithms in searching for the failure events. Herein, the pedestrians were controlled by AST to identify paths where the AV was likely to hit or collide with them. The experiments showed that DRL-based AST outperformed the MCTS-based scheme in finding failure scenarios. However, Corso \etal \cite{ast4av_rewardAug2019} pointed out that the results were uninteresting because most failures were caused by pedestrians walking into an already-stopped AV. To address this issue, they proposed incorporating the notion of responsibility-sensitive safety (RSS) \cite{rss2017} into the reward functions, and used the Trust Region Policy Optimization (TRPO) \cite{trpo} to solve the DRL problem. They found a more diverse range of failures showing the scenarios in which the IDM driven AV was acting improperly. 

Deep learning algorithms has become a key factor in supporting the AST to find scenarios of interest more efficiently. Underpinned by this approach, recent works \cite{ast4av_simulation2019,ast4av_fidelity2021} adopted Recurrent Neural Networks (RNN) \cite{rnn} and Long Short-Term Memory (LSTM) \cite{lstm} as alternatives to Multilayer Perception (MLP) networks to solve the RL problem. In particular, Mark \etal \cite{ast4av_fidelity2021} proposed the backward algorithm to deal with fidelity-dependents in simulation when applying AST to find failure scenarios. In their work, Proximal Policy Optimization (PPO) \cite{ppo2017} was used as the DRL solver, and NVIDIA's CarSim simulation was employed to demonstrate the testing effectiveness. Reuel \etal \cite{ast4av_ethic2022} further improved the previous study \cite{ast4av2018} by redefining the action space of AST to consider ethical dilemmas in autonomous systems, including ``do-not-harm'' pedestrians and ``do-not-harm'' occupants for the same AV-pedestrian collision avoidance scenario. Furthermore, Peter and Katherine \cite{ast4av_laneChange2021} studied the impact of human-in-the-loop and came up with an improved reward function that incorporates a probability of classifying a critical state from a neural network trained on annotated data. In addition, recent adversarial testing \cite{rel_dense_rl} introduced a dense deep reinforcement learning that amends the Markov decision processes by removing non safety critical states and reconnecting critical ones to apply to complex driving environments.

Table \ref{table:compare_related} summarizes the works that are related to our study. The key limitation with most existing studies is that their intelligent driving model used under test are very simple \cite{ast4av2018, ast_formulation2020, ast4av_ethic2022, ast4av_fidelity2021, ast4av_rewardAug2019, ast4av_simulation2019}. Recent attempts considered more complex systems driven by a DQN model \cite{highway-env2018, ast4av_laneChange2021}, which later showed an significant enhancement in the results. Another main limitation of existing AST studies is that they considered only longitudinal movement. Complex environments require more diverse actions, especially the lane change in the lateral direction. Furthermore, these framework tests with a limited number of traffic participants, which commonly only have a single pedestrian on a single-lane road. Another method, based on Reinforcement Learning (RL) \cite{ast4av_laneChange2021}, was used to create the driving model, which then required ground truth data for training the AST model may not always be available.

To address the aforementioned gaps, we propose and implement a new AST framework for testing AVs on multi-lane roads in this paper. In our framework, we develop a new reward function that takes into account both the collision of the ego AV (or system under test, SUT) and the safety of the surrounding vehicles, encouraging safe vehicle movements while closely maneuvering around the SUT to increase the likelihood of detecting potential crash scenarios. To demonstrate the effectiveness of our proposed framework, we develop and introduce an advanced driving model that builds on the existing Intelligent Back-looking Driving Model (IBDM) and Minimizing Overall Braking Induced by Lane Changes (MOBIL). Our driving model improves the flexibility and realism of the system under test (SUT), allowing it to perform a broader range of actions, such as safely navigating the multi-lane road, changing lanes, and avoiding collisions.

\subsection{Preliminaries of Reinforcement Learning (RL)}

In a reinforcement learning (RL) situation, the agent receives information about the state of the environment, denoted as $\mathrm{s}_t$, at each time step $t$. The agent then selects an action, denoted as $\mathrm{a}_t$, and based on this action, it receives a reward, denoted as $R_t$, and an updated state, denoted as $\mathrm{s}_{t+1}$, at time $t+1$. The objective of a RL agent is to acquire an optimal strategy $\pi^*:\mathbf{S}\rightarrow \mathbf{A}$, mapping from state spaces to action spaces which maximizes the cumulative reward $\mathrm{r}_t=\sum_{k=0}^{T}\gamma^k R_{t+k}$ where $R_{t+k}$ represents the reward at time step $t+k$, and $\gamma\in(0,1]$ is the discount factor, which quantifies the proportional relevance for prospective rewards. The state-action value function under policy $\pi$, represented as $Q^\pi(\mathrm{s}_t,\mathrm{a}_t)$, provides an estimate of the expected return, which is the total accumulated reward in an infinite time horizon. This estimate is based on starting from state $\mathrm{s}_t$, taking action immediately $\mathrm{a}_t$, and then following policy $\pi$ thereafter. The optimal Q-function is defined by the Bellman equation \cite{bellman1954theory}:
\begin{equation}
\begin{split}
    Q^*(\mathrm{s}_t,\mathrm{a}_t) &=\mathbb{E}[R(\mathrm{s}_t,\mathrm{a}_t)\\ &+\gamma\sum_{\mathrm{s}_{t+1}} P(\mathrm{s}_{t+1}|\mathrm{s}_t,\mathrm{a}_t)\text{max}_{\mathrm{a}_{t+1}}Q^*(\mathrm{s}_{t+1},\mathrm{a}_{t+1})]
\end{split}
\end{equation}
where the subsequent state $\mathrm{s}_{t+1}$ is obtained by sampling from the transition rules $P(\mathrm{s}_{t+1}|\mathrm{s}_t,\mathrm{a}_t)$ of the environment. The state value function of a state $\mathrm{s}_t$ under policy $\pi$, denoted as $V^\pi(\mathrm{s}_t)$, represents the expected cumulative reward when starting from state $\mathrm{s}_t$ and consistently following policy $\pi$, \ie, $V^\pi(\mathrm{s}_t=\mathbb{E}_\pi[\mathrm{r}_t|\mathrm{s}_t=\mathrm{s}]$. Frequently, the policy of the agent is defined by certain parameters $\theta$, and the objective is to acquire the suitable $\theta$ values in order to get the intended behavior of the system. In actor-critic algorithms, two networks are used: a critic network, parameterized by $\theta_c$, to learn the value function $V^{\pi_\theta}_{\theta_c}(\mathrm{s}_t)$, and an actor network, $\pi_{\theta_a}(\mathrm{a}_t|\mathrm{s}_t)$, parameterized by $\theta_a$. The policy network is updated by maximizing the given objective function:
\begin{equation}
    J^{\pi_{\theta_a}}=\mathbb{E}_{\pi_{\theta_a}}[\log \pi_{\theta_a} (\mathrm{a}_t|\mathrm{s}_t) U_t]
\end{equation}
where $U_t=Q^{\pi_\theta}(\mathrm{s}_t,\mathrm{a}_t)-V_{\theta_c}(\mathrm{s}_t)$ is the function that quantifies the increase in reward achieved by taking action $\mathrm{a}_t$ compared to the average reward obtained from all potential actions done at state $\mathrm{s}_t$. The value function parameter $\theta_c$ is updated through the process of minimizing the loss function provided below:
\begin{equation}
    J^{V_{\theta_c}}=\min_{\theta_c} \mathbb{E}_{\mathcal{D}} \left(R_t+\gamma V_{\theta'_c}(\mathrm{s}_{t+1})-V_{\theta}(\mathrm{s}_t) \right)^2
\end{equation}
where $\mathcal{D}$ refers to an experience replay buffer that stores past experiences. The symbol $\theta'_c$ represents the parameters obtained from previous iterations, which are employed in a target network.

In the following sections, we will describe and discuss our proposed framework in details.
\section{Proposed Adaptive Stress Testing (AST) Framework}\label{section:method}
We propose in this section our novel framework to identify critical scenarios in which an autonomous vehicle (or ego vehicle) may crash while driving on multiple lanes road under complex traffic conditions. In this environment, all vehicles move with high speeds, and perform a wide range of driving maneuvers including lane changing and acceleration/deceleration. We take into account these behaviours to construct our new framework and utilize a reinforcement learning model to support the AST in identifying scenarios in which the ego vehicle might collide with other vehicles on the multi-lane road. To achieve this, we have formulated the problem of finding failure events as a sequential process of decisions following the work of Koren \etal \cite{ast_formulation2020}.

\begin{figure}[htbp]\centering
\includegraphics[width=0.45\textwidth]{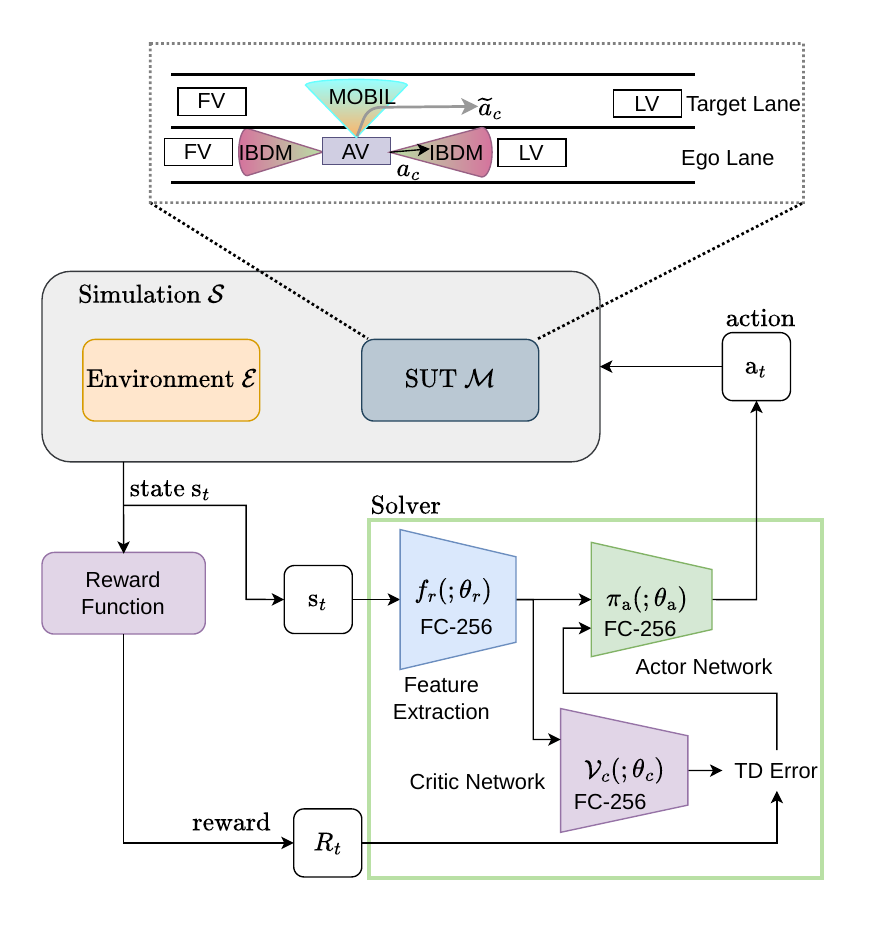}%
    \caption{Our proposed Adaptive Stress Testing framework for finding crashed scenario in the multi-lane roads. AV, FL and LV abbreviations refer to the vehicle (\ie, ego vehicle), following vehicle, and leading vehicle, respectively.}
    \label{fig:ast_overall}
\end{figure}

The design of our framework comprises three primary components, that is, the simulator $\mathcal{S}$, the reward function $R$, and the solver. 
The simulator evolves through different stages as the vehicle and other agents in the simulation environment interact. As the simulator transitions to a new state, it communicates any pertinent information regarding the state to the module associated with the AST reward function to obtain a reward value. Subsequently, the calculated reward is passed to the solver where the policy is updated, enabling it to generate the next environmental action. The overall architecture is presented in Figure \ref{fig:ast_overall} with the details being described below.

\subsection{The simulator}
The simulator $\mathcal{S}$ consists of an intelligent model for an autonomous vehicle $\mathcal{M}$ being used as a system under test (SUT) and the environment $\mathcal{E}$. This intelligent driving model ensures safe movements of the AV in the environment. We further define a set of goal states $E$ for the simulator, which represents the possible failure of the SUT. In this study, $E$ consists of all states that the ego vehicle likely to collide with other traffic participants.
The goal of the RL solver is to update the state of the simulator $\mathcal{S}$ by generating environment actions that comply with the driving model and road rules while exploring the corner scenarios where collision is likely. 

The AST technique is capable of managing a black-box or semi-black-box simulator that does not require the revelation of its complete internal states, and follows the following procedures:

\begin{itemize}
    \item \textsc{Init}$(\mathcal{S})$: Reset the simulation to its starting state after the previous simulation epoch has terminated.
    \item \textsc{Step}$(\mathcal{S}, \mathrm{a})$: Perform an action $\mathrm{a}_t$ at the time step $t$ and advance the simulator in time. Evaluate whether the new state of the simulation lies in the set of goal states $E$ and produce an indicator of this.
    \item \textsc{IsTerminal}$(\mathcal{S})$: Determine whether the current state belongs to the set of goal states $E$, or if the specified time limit has been exceeded. 
\end{itemize}

The state of environment consists of all three vectors associated with the actions of the ego vehicle, environment actions by vehicles other than the ego vehicle, and the observations, that is,
\begin{equation}
\mathrm{s}_{env} = \left( \mathrm{a}_{ego}, \mathrm{a}_{env}, \mathrm{o}_{env}\right)
\end{equation}
in which
\begin{itemize}
\item
$\mathrm{a}_{ego}$ is the action of the ego vehicle which include $a^{ego}_x$ and $a^{ego}_y$ corresponding to the components of acceleration/deceleration in the longitudinal and latitudinal directions, respectively. 
\item 
$\mathrm{a}_{env}=\left(\mathrm{a}^{(1)}_{env},\mathrm{a}^{(2)}_{env},\ldots,\mathrm{a}^{(n)}_{env}\right)$, is the environment action at each time step associated with $n$ surrounding vehicles (other than the ego vehicle), and $\mathrm{a}^{(i)}_{env}$ is the high-level action for the $i$-th closest vehicle $(i=1,2,\ldots,n)$ that each vehicle's action $\mathrm{a}^{(i)}_{env}$ can be one of several possible maneuvers, namely, left lane change, right lane change, acceleration, deceleration, or maintaining a constant speed. 
\item 
$\mathrm{o}_{env} = \left(\mathrm{o}^{(1)}_{env},\mathrm{o}^{(2)}_{env},\ldots,\mathrm{o}^{(n)}_{env}\right)$
is the observation that provides the kinematic information of both the ego vehicle and surrounding vehicles, where $\mathrm{o}^{(i)}_{env}$ represents the kinematic information of the $i$-th surrounding vehicle, which includes its positions $x^{(i)}$ and $y^{(i)}$, as well as its velocities $v_x^{(i)}$ and $v_y^{(i)}$, respectively, that is, $\mathrm{o}^{(i)}_{env} = (x^{(i)},y^{(i)},v_x^{(i)},v_y^{(i)})$ $(i=1,2,\ldots,n)$. Here, $v_x^{(i)}$ and $v_y^{(i)}$ (and $x^{(i)}$, $y^{(i)}$) refer to the longitudinal and latitudinal components, respectively, of the relative velocity (and relative position) between the ego vehicle and the $i$-th surrounding vehicle.
\end{itemize}



\subsection{Reward function}
We propose a novel reward function that is specifically designed to guide the simulation to identify scenarios in which a collision might occur in a multi-lane road environment. The proposed reward function overcomes the limitations in previous works, where only longitudinal trajectory is considered. More specifically, the primary limitation of existing rewards is the issue of sparse rewards. The multi-agent is only penalized if a collision is discovered when the state ends, \ie, at the end of which could be a long trajectory which leads to difficulty in learning how to find new cases with collisions. 

Herein, we introduce a new reward function which further accounts for collisions along both longitudinal and lateral direction. Its aim is to to encourage the surrounding vehicle to move closely to the ego vehicle on the multi-lane road both horizontally and vertically, leading to a more diverse range of actions and potentially higher likelihood of a collision. To measure the collision probability of other vehicles against the ego vehicle, we utilize the notion of time-to-collision (TTC) \cite{ttc_2013}. In our study, at each time step, the multi-lane road environment may consist of up to six vehicles moving around the ego vehicle. They include the ones in front and rear, spanning three lanes: the ego lane, the left lane, and the right lane. The collision measurement of ego vehicle is calculated based on the collision probability of the ego vehicle with surrounding vehicles via comparing their TTC value with a predefined time threshold. Similar to the theory of using TTC time to determine cutting-in and cutting-out scenarios as proposed by Euro NCAP \cite{euro2019assessment,euro_ncap_2023}, we evaluate if the TTC time is less than the given threshold, the situation is then considered a collision where the collision probability is set to 1. Otherwise, the collision probability of a vehicle is the ratio between the TTC time of that vehicle to the ego vehicle and the TTC threshold shown in Equation \ref{eq:ego_ttc}. Finally, we calculate the ego vehicle collision probability by aggregating the collision probabilities of all surrounding vehicles into an overall score in Equation \ref{eq:ego_collision}.

\begin{figure}[htbp]
    \centering
\includegraphics[width=0.45\textwidth]{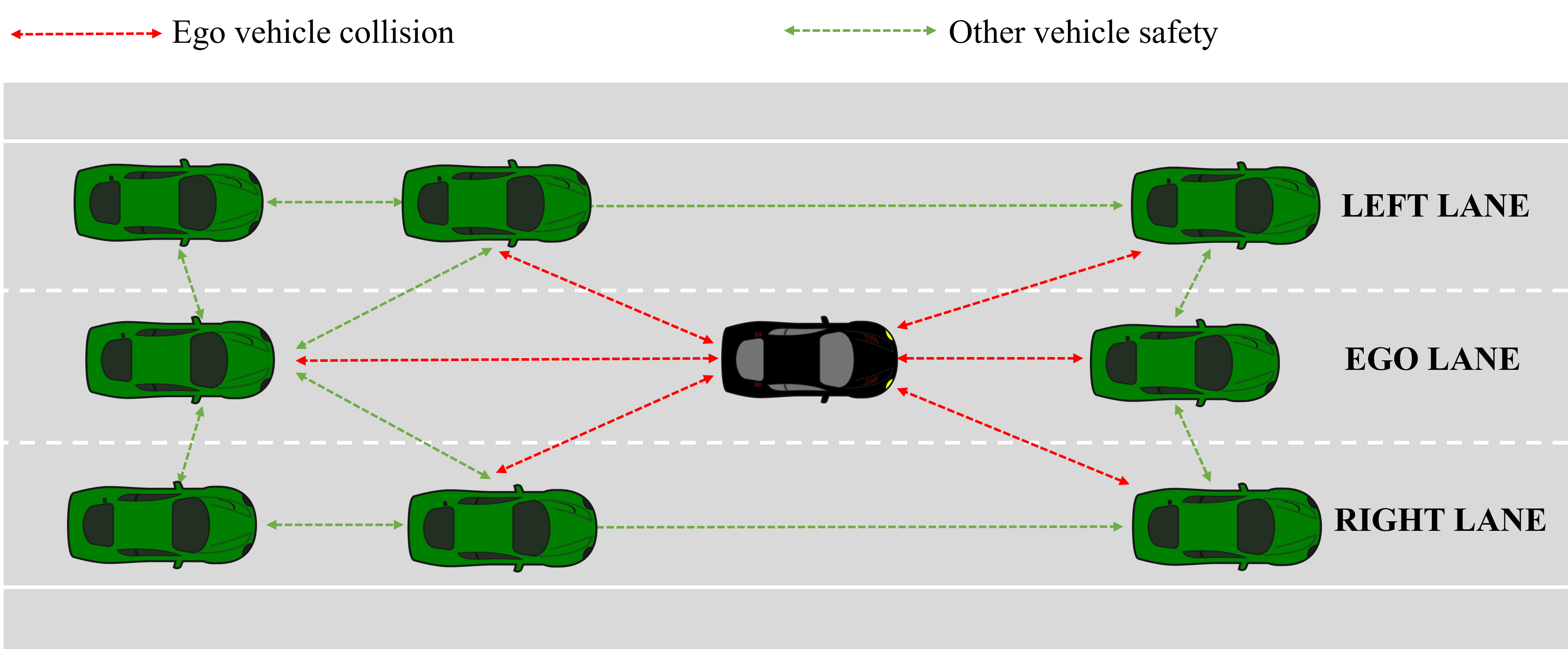}
    \caption{Illustration of interaction between vehicles with two components: other vehicle safety and ego vehicle collision.}
    \label{fig:reward}
\end{figure}

Our reward function is defined based on two quantities: (i) the measurement of collision with the ego vehicle, being denoted as $\Phi_{\text{coll}}$, and (ii) the measurement of safety of vehicles surrounding the ego vehicle against other vehicles, being denoted as $\Psi_{\text{safe}}$. Figure~\ref{fig:reward} illustrates this two components and their impact on the movements of the ego vehicle and surrounding vehicles. The proposed reward function is defined as follows:
\begin{numcases}{R({\mathrm{s}_t}) =}
    0 & \hspace{-1em} $\mathrm{s}_t \in E$ \label{eq:case_1} \\
    -\alpha - \beta \min\limits_{\mathrm{v}_q \in \mathbf{N}(\mathrm{v}_e)} t_{\text{TTC}}(\mathrm{v}_e, \mathrm{v}_q) & \hspace{-1em} $\mathrm{s}_t \notin E, t\geq T$ \label{eq:case_2} \\
    \lambda \Phi_{\text{coll}}(\mathrm{s}_t) + (1-\lambda)\Psi_{\text{safe}}(\mathrm{s}_t) & \hspace{-1em} $\mathrm{s}_t \notin E, t < T$ \label{eq:case_3}
\end{numcases}
where $E$ is set of failure states (or collision scenarios) which is determined by intersecting two polygons covering vehicles similar to previous study \cite{highway-env2018}; $\mathbf{N}(\mathrm{v}_e)$ is the list of surrounding vehicles close to the ego vehicle $\mathrm{v}_e$ and $\mathbf{N}(\mathrm{v}_i)$ is the set of surrounding vehicle close to the given vehicle $\mathrm{v}_i$; $t_{TTC}(\mathrm{v}_p,\mathrm{v}_q)$ is the time-to-collision between two vehicles $\mathrm{v}_p$ and $\mathrm{v}_q$, calculated based on Saffarzadeh \etal \cite{ttc_2013}; and $\alpha$ and $\beta$ in Equation \ref{eq:case_2} are parameters to penalize for failing to identify a collision \cite{ast4av2018,ast4av_rewardAug2019}. As we observe the maximum number of six vehicles closest to our ego vehicle mentioned earlier, the length of $\mathbf{N}(\mathrm{v}_e)$ is less than or equal to 6.

We define a formulation as in the following equation to support the calculation of collision and safety probabilities later on.
\begin{equation}
    \mathbb{E}_{\mathrm{v}_j\sim \mathbf{N}(\mathrm{v}_i)}\left[\zeta\right] = \frac{1}{\lVert \mathbf{N}(\mathrm{v}_i) \rVert}\sum _{\mathrm{v}_j \in \mathbf{N}(\mathrm{v}_i)} \zeta(; \mathrm{v}_j)
\end{equation}
where $\zeta$ is a function.

In our proposed reward function, we have introduced a weighting parameter $\lambda$ which has a value in the range of $[0, 1]$ in Equation \ref{eq:case_3} for balancing between the safety measurement of surrounding vehicles (\ie, $\Psi$) and the collision measurement of the ego vehicle (\ie, $\Phi$). As explained earlier, the collision measurement of the ego vehicle can be expressed as:
\begin{equation}\label{eq:ego_collision}
    \Phi_{\text{coll}}(\mathrm{s}_t) = \log \mathbb{E}_{\mathrm{v}_p\sim \mathbf{N}(\mathrm{v}_e)}\left[P_{\text{coll}}(\mathrm{s}_t, \mathrm{v}_e)\right]
\end{equation}
where the collision probability $P_{\text{coll}}(\mathrm{s}_t, \mathrm{v}_p, \mathrm{v}_q)$ of two vehicles $\mathrm{v}_p$ and $\mathrm{v}_q$ at a state $\mathrm{s}_t$ is calculated as below:
\begin{equation}\label{eq:ego_ttc}
\begin{aligned}
P_{\text{coll}}(\mathrm{s}_t, \mathrm{v}_p, \mathrm{v}_q) = 
    \begin{cases}
     1 & \hspace{-3em} \text{if } t_{TTC}(\mathrm{v}_p, \mathrm{v}_q) \leq t_{\Delta} \\
     \frac{t_{\Delta}}{t_{TTC}(\mathrm{v}_p,\mathrm{v}_q)} & \text{otherwise}
    \end{cases}
\end{aligned}
\end{equation}
Here, $t_{\Delta}$ is the time threshold where two vehicles are too close to each other and likely to collide.

At the same time, we are encouraging the surrounding vehicles to move safely without colliding with other vehicles. The measurement of safety movement of ego vehicle against other vehicles is given by
\begin{equation}
    \Psi(\mathrm{s}_t) = \mathbb{E}_{\mathrm{v}_q\sim \mathbf{N}(\mathrm{v}_e)}\left[\Theta(\mathrm{s}_t)\right]
\end{equation}
where $\mathbf{N}_{\text{surr}}(\mathrm{v}_i)$ denotes the list of vehicles surrounding a given vehicle $\mathrm{v}_i$ within a distance $d$ of the vehicle $\mathrm{v}_i$ in any direction.
The function $\Theta(\mathrm{s}_t, \mathrm{v}_i)$ represent the safety probability of a given vehicle $\mathrm{v}_i$ at the state $\mathrm{s}_t$, as below:
\begin{equation}
    \Theta(\mathrm{s}_t, \mathrm{v}_i) = \log \mathbb{E}_{\mathrm{v}_j\sim \mathbf{N}(\mathrm{v}_i)}\left[P_{\text{safe}}(\mathrm{s}_t, \mathrm{v}_i)\right]
\end{equation}
where:
\begin{equation} \label{eq:safe_ttc}
P_{\text{safe}}(\mathrm{s}_t,\mathrm{v}_p,\mathrm{v}_q) = 
    \begin{cases}
        0 & \hspace{-3em} \text{if } t_{TTC}(\mathrm{v}_p,\mathrm{v}_q) \leq t_{\Delta} \\
        1 - \frac{t_{\Delta}}{t_{TTC}(\mathrm{v}_p,\mathrm{v}_q)} & \text{otherwise}\\
    \end{cases} 
\end{equation}
We set the safety probability to be minimum, \ie, 0, when the TTC time is less than a threshold. We use the same threshold value $t_{\Delta}$ to determine both the safety probability and the collision probability.

\subsection{The solver} 
We adopt Deep Q-Learning approach for the solver. Following the work of Raffin \etal \cite{stable-baselines3}, our network architecture is composed of three components, namely, the Feature extraction $f_r(;\theta_r)$, the Critic network $\mathcal{V}_c(\mathrm{s}_t;\theta_{c})$, and the Actor network $\pi_\mathrm{a}(;\theta_\mathrm{a})$. Its core component, \ie, the Feature extraction $f_r$ being parameterized by $\theta_r$, helps extract features from the environment state $\mathrm{s}_t$. 
In deep Q-learning, the function approximation can result in errors due to overvalued estimations. To address the issue, we parameterized critic network $\mathcal{V}_c$ by $\theta_c$ and estimate the error associated with Twin Delayed Deep Deterministic policy gradient (TD error) by taking the reward value $R_t$ and extracted features as input as in Fujimoto \etal \cite{td3_2018}. Following \cite{actor_critic}, the TD error is then used by the Actor network $\pi_a$ along with the extracted features from $f_r$ to approximate the environment action $\mathrm{a}_t$ as follows:
\begin{equation}
    \text{TD Error} = R_{t+1}+\gamma \mathcal{V}_c(\mathrm{s}_{t+1};\theta_c)-\mathcal{V}_c(\mathrm{s}_t;\theta_c)
\end{equation}
where $\gamma$ is a discount factor ($0<\gamma<1$).

The deep neural network is composed of fully connected (FC) layers consisting of 256 neurons as shown in the figure \ref{fig:ast_overall}. The function $f_r$ will extract features from the state data of the environment $\mathrm{s}_t$. This feature will be used by both the actor and critic networks. The critic network's output is the state value, which will be combined with the reward function to calculate the TD error. The actor network produces five action probabilities, which correspond to the five possible maneuvers of the vehicle described in Sec \ref{subsec:veh_action} below.

\subsection{Intelligent Driving Model (uIDM)}
To facilitate the testing we propose a comprehensive intelligent driving model referred to as the unified Intelligent Driving Model (uIDM) which accounts for both complex longitudinal and latitudinal movements on multi-lane roads. The proposed uIDM departs from all the System Under Test (SUT) used in current AST framework testing that only consider the longitudinal direction, which makes it an unrealistic driving maneuver for AVs.

The proposed uIDM model is a combination of the Intelligent Back-looking Driving Model (IBDM) and MOBIL models both of which have been used in the literature to simulate driving behaviors and study the safety of vehicles in front and sides but not at the rear of the ego vehicle. To address such limitations we build the uIDM based on the combination of IBDM and MOBIL to control the ego vehicle's safe movement with both sides, front and rear vehicles \cite{ibdm2020}. The new intelligent driving model can handle more sophisticated tasks in the multi-lane road environment due to its action capability as described below.

\subsubsection{Longitudinal actions}
uIDM is based on the assumption that vehicles on the multi-lane road are connected and thus information about all the front and rear vehicles surrounding the ego vehicle is available in real time. As a result, each vehicle's acceleration might be adjusted to improve driver behavior and increase stability.
The equations for determine the longitudinal acceleration or deceleration of a vehicle is expressed as follows: 
\begin{equation}
a^{(i)}(t) = a^{(i)}_{max}\left( a^{(i)}_{acc}(t) + \left(a^{(i)}_{dec}(t) \right)^2 \right ) 
\end{equation}
where
\begin{equation}
a^{(i)}_{acc}(t) = 1 - (\frac{v_e(t)}{v_0})^{\delta}
\end{equation}

\begin{equation}
a^{(i)}_{dec}(t) = \frac{s^*\left(v_e(t), \Delta_{v_{e,l}(t)}\right )}{s_e(t)} - \epsilon\frac{s^*\left(v_f(t), \Delta_{v_{f,e}(t)} \right )}{s_f(t)}
\end{equation}
in which
\begin{align}
\Delta_{v_{f,l}(t)} &= v_f(t) - v_l(t)\\
s^*\left(v_f(t),\Delta_{v_{f,l}(t)}\right) 
 &= s_0 + \mu v_f(t) + \frac{v_f(t)\Delta_{v_{f,e}(t)}}{2\sqrt{|a^{(i)}_{max} a^{(i)}_{min}|}}
\end{align}
In these equations, $v_0$ is the desired velocity; $s_0$ represents the minimal spatial separation between two neighboring vehicles in a highly congested traffic scenario; $\mu$ is desired time gap; $v_e(t), v_f(t), v_l(t)$ are the velocity of ego, following, leading vehicle respectively; $a^{(i)}_{max}$, $a^{(i)}_{min}$ are the maximum acceleration and deceleration, respectively; $\epsilon$ is the hyperparameter; and $\delta$ is the acceleration exponent that quantifies the degree to which the acceleration is influenced by the speed and gap.

\begin{figure}[tbp]
    \centering
    \includegraphics[width=0.45\textwidth]{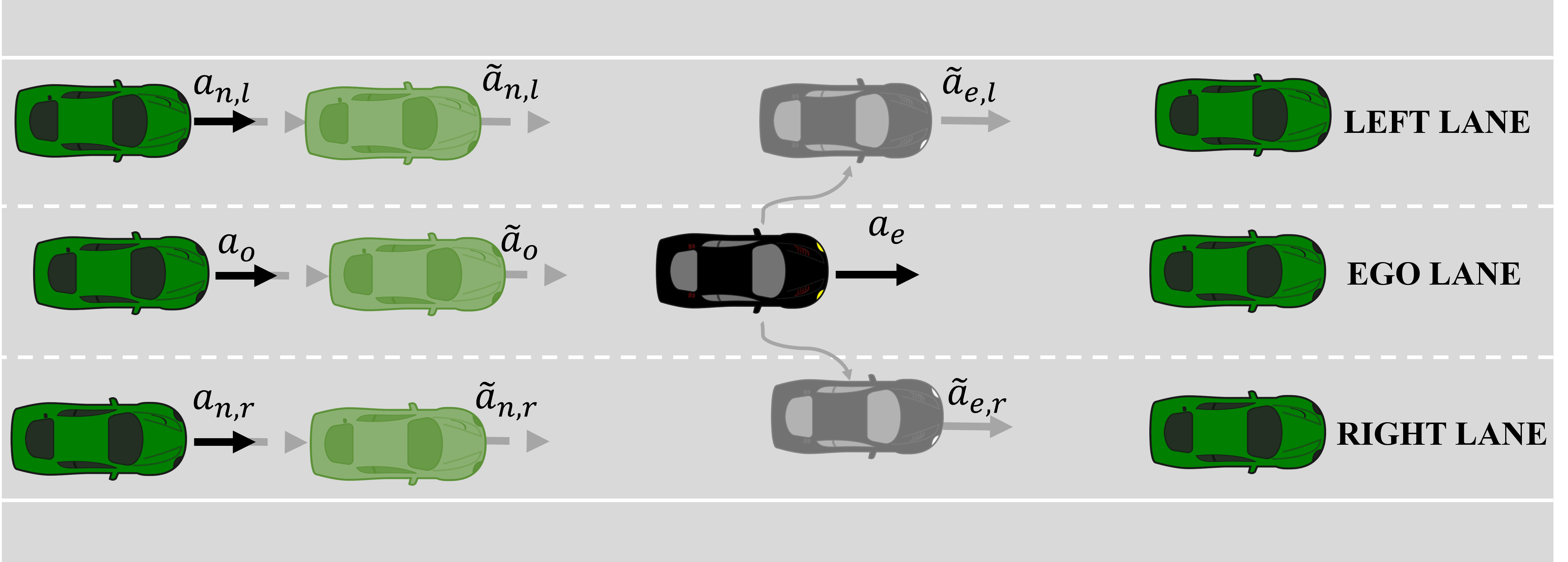}
    \caption{Lane change scenario with our uIDM which is similar to MOBIL \cite{mobil2007}. The following vehicles in the left, ego, and right lanes each have their own accelerations, denoted as $a_{n,l},a_{o},a_{n,r}$ respectively. The ego vehicle, which is currently accelerating at a rate of $a_e$, has the option to switch to either the left lane or the right lane, with new accelerations of $\widetilde{a}_{e,l}$ and $\widetilde{a}_{e,r}$, respectively. After the ego vehicle changes lanes, the following vehicle can move with the new accelerations $\widetilde{a}_{n,l},\widetilde{a}_o,\widetilde{a}_{n,r}$, respectively.}
    \label{fig:mobil}
\end{figure}

\subsubsection{Lateral actions}
The MOBIL model is integrated to determine how safe it is to change lanes by considering the speeds of surrounding vehicles relative to the ego vehicle. Figure~\ref{fig:mobil} illustrates how this algorithm considering dynamics of the vehicles to make the lane changing.
Same as MOBIL, the decision-making process of uIDM has two steps. First, it makes sure that when the lane change happens the car behind it does not slow down too much which is formulated by the equation:
\begin{equation}
    \widetilde{a}_n > a_{safe}
\end{equation}
where $\widetilde{a}_n$ is the following (left or right) vehicle's acceleration after changing lanes, and $a_{safe}$ is the ceiling threshold of acceleration when vehicle braking. Second, if the first safety criterion is met, uIDM continues to check the below condition before deciding to change lane \cite{mobil2007}:
\begin{equation}\underbrace{\widetilde{a}_e-a_e}_\text{driver} + \rho\left (\underbrace{\widetilde{a}_n-a_n}_\text{next follower} + \underbrace{\widetilde{a}_o-a_o}_\text{previous follower} \right ) \geq \Delta a_{th}
\label{eq:mobil}
\end{equation}
where subscript $e$ denotes ego vehicle, while $o$ and $n$ stand for the previous and next followers after the lane change. The variable $a$ represents the vehicle's acceleration before changing lane, and the variable $\widetilde{a}$ denote the vehicle's acceleration after changing the lane. In the lane change action, the value of the variable $\rho$ is a politeness coefficient, and $\Delta a_{th}$ represents the acceleration gain required for performing a lane change.


\section{Experiment}\label{section:exp}
\subsection{System setup}
We describe the environmental setup and ways to evaluate the performance of our proposed approach. 
The multi-lane road environment is used to test the driving model, which consists multiple lanes where multiple vehicles move at different speeds. The \textsc{autonomous driving simulator} \cite{highway-env2018} package, introduced by OpenAI Gym, is used to train a vehicle controller whose goal is to gain a high speed while avoiding collisions with surrounding vehicles.
We use the proximal policy optimization (PPO) algorithm \cite{ppo2017} from Stable Baseline3 \cite{stable-baselines3} to train multiple agents and optimize the reinforcement learning policy.

\subsection{Parameters}
To penalize the algorithm for failing to find a collision, the constants $\alpha$ and $\beta$ are set to 10,000 and 1000, respectively, similar to previous works \cite{ast4av_rewardAug2019,ast4av2018}. Regarding surrounding vehicles, distance $d$ is set to 100m.
For uIDM, we set $\epsilon$ to 0.4, $\mu$, $s_0$, $\delta$, and $a_{max}^i, a_{min}^i$ to 1.5(s), 10(m), 4.0, 3.0(m/s2), and -5.0(m/s2) in longitudinal control, $a_{safe}$ to -2.0(m/s2), $\rho$ to 0.0, and $\Delta a_{th}$ to 0.2(m/s2) for lateral control.
In our experiment, 40 vehicles are driving on a four-lane road. The speed limit on the road is 25 m/s (90 km/h). The vehicles are given a random starting speed of 25 m/s. The time-to-collision thresholds $t_{\Delta}$ are 1.5, 2, 2.5, 3, 3.5 seconds similar to the time-to-collision suggested by the Euro NCAP \cite{euro2019assessment} to determine dangerous scenarios such as cutting-in, cutting out. The scenario runs for $4\times10^4$ time steps. Our experiment is run on a machine os Ubuntu 20.04 and equipped with an NVIDIA GeForce P8 4 GB GPU and 24 GB RAM.
The RL was trained with a batch size of 32, $T$ of 500 time steps, a discount factor $\gamma$ of 0.8, and a learning rate of $5\times 10^{-4}$. The neural network policy is composed of up of two hidden layers, each with 256 neural units. Code and supplementary materials are publicly available in our repository\footnote{https://github.com/khaclinh/AST4AV-HighWay}.

\subsection{Calibration}
In our proposed reward function, the trade-off between two components (\ie, the safety and collision measurements) is controlled primarily by the variable $\lambda$. To determine the appropriate $\lambda$, we design a series of experiments where $\lambda$ takes one of the values in the set \{0, 0.1, 0.2, 0.3, 0.4, 0.5, 0.6, 0.7, 0.8, 0.9, 1\}. Then, the statistics of California Autonomous Vehicles crashes \cite{crash_stat2021} (CC) is used as a ground truth to determine the value of $\lambda$ that best represents the crashes associated with autonomous vehicles in reality. Herein, we divide the types of crashes discovered by our framework into three groups: lane-change collisions, rear-end collisions, and other types.

\begin{table}[tbp]
\caption{Comparison of our framework performance with different configurations of $\lambda$ against the observations of crashes in California \cite{crash_stat2021} (CC).
Bold value indicate the closest value to observed value of the crash data.}
\label{table:statistic_realistic}
\begin{center}
        \begin{tabular}{|c|rrr|r|} \hline
        \textbf{Parameter} & \textbf{Rear-end} & \textbf{Lane change} & \textbf{Other} & \textbf{$\mathcal{D}_e$} \\ 
        \hline\hline 
        $\lambda=0.0$   & 70.63\%             & 9.54\%                 & 19.83\%          & 24.87\%            \\ \hline
        $\lambda=0.1$   & 65.10\%              & 10.53\%                & 24.37\%          & 20.61\%            \\ \hline
        $\lambda=0.2$   & 67.33\%             & 20.07\%                & 12.60\%           & 18.27\%            \\ \hline
        $\lambda=0.3$   & 65.97\%             & 16.62\%                & 17.41\%          & 17.12\%            \\ \hline
        $\lambda=0.4$   & 60.62\%             & 20.79\%                & 18.59\%          & 10.25\%            \\ \hline
        $\lambda=0.5$   & 58.14\%             & \textbf{21.82\%}                & \textbf{20.04\%}          & 7.41\%            \\ \hline
        $\lambda=0.6$   & 60.53\%             & 19.89\%                & 19.58\%          & 10.52\%            \\ \hline
        $\lambda=0.7$   & 58.87\%             & 20.18\%                & 20.95\%          & 8.98\%            \\ \hline
        $\lambda=0.8$   & \textbf{56.21\%}             & 21.90\%                 & 21.89\%          & \textbf{5.97\%}            \\ \hline
        $\lambda=0.9$   & 57.97\%             & 21.23\%                & 20.80\%           & 7.61\%            \\ \hline
        $\lambda=1.0$   & 58.56\%             & 20.05\%                & 21.39\%          & 8.86\%           \\ \hline\hline
        \underline{CC}     & \underline{52.46\%}             & \underline{26.47\% }               & \underline{20.07\%}          &  -                 \\ \hline        
        \end{tabular}
\end{center}

\end{table}
Table \ref{table:statistic_realistic} shows the detailed results of our frameworks' collision group statistics in comparison with the California crash data. The results demonstrated that, with each collision type (rear-end, lane change, or other), our framework using $\lambda=0.8$ has a percentage of rear-end caused collisions similar to that of observed crash data, and with $\lambda=0.5$, the percentage of lane-change caused collisions and other types is similar between the model and observed data, respectively. We measure the similarity of these crash types using Euclidean distance to better determine the most appropriate $\lambda$ in overall collisions. The Euclidean distance $\mathcal{D}_e(\mathrm{q}^{\lambda},\mathrm{q}^{CC})$ is defined as below:
\begin{equation}
    \label{eq:eu_distance}
    \mathcal{D}_e = \sqrt{(\mathrm{q}^{\lambda}_r - \mathrm{q}^{CC}_r)^2 + (\mathrm{q}^{\lambda}_l - \mathrm{q}^{CC}_l)^2 + (\mathrm{q}^{\lambda}_o - \mathrm{q}^{CC}_o)^2}
\end{equation}
where $\mathrm{q}^{CC}_r, \mathrm{q}^{CC}_l, \mathrm{q}^{CC}_o$, and $\mathrm{q}^{\lambda}_r, \mathrm{q}^{\lambda}_l, \mathrm{q}^{\lambda}_o$ are the statistics of rear-end, lane-change, other types of the California crash data, and the corresponding values obtained from our method using different value of $\lambda$ as shown in Table \ref{table:statistic_realistic}.
The Euclidean distances are also included in Table \ref{table:statistic_realistic}. Observe that $\lambda=0.8$ gives a smallest $\mathcal{D}_e$ distance and thus will be used in our experiments reported in the next section.

\section{Results and Discussions}\label{section:result}
\subsection{Results}
\begin{table*}[t]
\caption{Statistics of collision scenarios for both ego and other vehicle maneuvers leading to collisions. {S, R, L are directions of ego vehicle going straight forward along ego lane, changing to the right lane, and shifting to the left lane, respectively. SV\_LLC, SV\_RLC, SV\_CSK, SV\_A, SV\_D are five behaviors of crashed vehicle: change to the left lane, change to the right lane, keep constant speed, accelerate and decelerate.}}
\label{table:stat_collision_vehicle_maneuver}
\centering
\begin{tabular}{|l|c|c|c|c|c|c|c|c|c|c|c|c|} \hline
\multirow{4}{*}{\textbf{Crashed vehicle}} & \multicolumn{12}{c|}{\textbf{System Under Test (SUT)}}\\ 
\cline{2-13}
& \multicolumn{3}{c|}{\textbf{AST4AV \cite{ast4av2018}}} & \multicolumn{9}{c|}{\textbf{Our study}} \\ 
\cline{5-13} & \multicolumn{3}{c|}{} & \multicolumn{3}{c|}{$\lambda=0.8$)} & \multicolumn{3}{c|}{$\lambda=1$} & \multicolumn{3}{c|}{$\lambda=0$} \\ \cline{2-13}
 & S     & R     & L    & S     & R     & L       & S     & R     & L    & S     & R     & L  \\ 
\hline\hline 
SV\_LLC    & 182          & 17        & 21   & 304        & 25      & 29   & 239            & 23          & 26         & 237         & 7        & 10          \\  \hline
SV\_RLC                             & 143          & 8         & 7       & 285        & 31      & 35   & 251            & 27          & 28         & 323         & 10       & 9        \\  \hline
SV\_CSK                          & 72           & 10        & 15     & 94        & 23      & 21     & 102            & 18          & 17         & 61          & 7        & 9       \\  \hline
SV\_A                                 & 139          & 13        & 18     & 131        & 22      & 31    & 102            & 14          & 17         & 45          & 3        & 9        \\  \hline
SV\_D                                  & 17           & 1         & 2     & 35         & 9       & 12    & 16             & 2           & 6          & 26          & 7        & 3        \\ \hline
\hline 
$\sum$ 1 action
      & 553          & 49        & 63      & 849        & 110     & 128    & 710            & 84          & 94         & 692         & 34       & 40     \\ \hline
$\sum$ all actions
          & \multicolumn{3}{c|}{665}          & \multicolumn{3}{c|}{1,087}     & \multicolumn{3}{c|}{888}                   & \multicolumn{3}{c|}{766}        \\ \hline
\end{tabular}
\end{table*}

We compare performance of our proposed framework with the state-of-the-art of Adaptive Stress Testing for Autonomous Vehicles (AST4AV) \cite{ast4av2018}. We use the same set of critical states $E$ in both reward functions. Both of our framework and AST4AV framework are used to test the proposed uIDM driving model.
 In particular, we analyze the performance obtained from the two frameworks when a collision occurs such as vehicle maneuvers, collision type, vehicle speeds, and vehicle changing lanes and along with analysis of our proposed reward function.

\subsubsection{Vehicle maneuvers} \label{subsec:veh_action}
First, we analyze the vehicle maneuvers when a collision occurs. In the collision statistics described in Table \ref{table:stat_collision_vehicle_maneuver}, we combine three ego vehicle behaviors (go straight, turn left, turn right) with five crashed vehicle behaviors (change to the left lane, change to the right lane, keep a constant speed, accelerate, and decelerate). Our framework identifies more crashed scenarios (1,087) as compared to AST4AV \cite{ast4av2018} (665 crashed scenarios) (Table \ref{table:stat_collision_vehicle_maneuver}). Furthermore, our framework identifies a larger and more diverse set of diverse crashed scenarios than AST4AV in all of the behaviors of both the ego and crashed vehicle. This result demonstrates that our framework outperforms the AST4AV \cite{ast4av2018} in identifying crashed scenarios.

To analyze the impact of the two components in our proposed framework on failure scenario discovery capabilities, we add to Table \ref{table:stat_collision_vehicle_maneuver} the results of our framework using $\lambda=0$ and $\lambda=1$, which correspond to the absence of collision measurement of ego vehicle, and safety measurement of surrounding vehicles, respectively. The results show that when one of these components is absent, the reward functions identify less crashed scenarios than when both are considered with $\lambda=0.8$ weighting.

\subsubsection{Collision type}
We categorize collisions that happen to the ego vehicles into six types based on the location of the crashed vehicle at the time step immediately preceding the collision with ego vehicle, which is in the front, rear, or in the left, right, and same lane of ego vehicle. Specifically, they are referred to as crashed vehicle leading and coming from the left lane (FL), crashed vehicle leading in the ego lane (FE), crashed vehicle leading and coming from the right lane (FR), crashed vehicle following and coming from the left lane (RL), crashed vehicle following in the ego lane (RE), and crashed vehicle following and coming from the right lane (RR).
\begin{table}[htbp]
    \begin{center}
    \caption{Statistics of collision types over different value of $\lambda$ of our framework in comparison with AST4AV \cite{ast4av2018}.}
    \label{table:comparison_collision_type}
    \footnotesize{
    \begin{tabular}{|l|c|c|c|c|c|c|c|} \hline
\multirow{2}{*}{\textbf{Framework}} & \multicolumn{7}{c|}{\textbf{Collision types}}                                                          \\ \cline{2-8}
     & FL & FE & FR & RL & RE & RR & Total \\ \hline\hline
AST4AV \cite{ast4av2018}             & 10           & 152           & 5           & 33           & 436          & 29           & 665             \\ \hline
   \textbf{Our study}, $\lambda=1$   & 14           & 258          & 17           & 43           & 520         & 36           & 888            \\ \hline
    \textbf{Our study}, $\lambda=0$         & 2           & 160          & 2           & 36           & 541         & 25           & 766            \\ \hline
    \textbf{Our study}, $\lambda=0.8$        & 19          & 296          & 30           & 60          & 611         & 71          & 1,087    \\ \hline       
    \end{tabular}
    }
    \end{center}

\end{table}

Table \ref{table:comparison_collision_type} presents the detailed collision types of AST4AV \cite{ast4av2018} and our framework with different components. Overall, both frameworks have a major number of collision in the RE or FE types. With AST4AV \cite{ast4av2018}, the identified collisions are heavily biased to the RE type while our framework clearly detects more the FE collision type.

\subsubsection{Other characteristics}
Below we study the effectiveness of our proposed framework. To this end, we conduct a comparative analysis between our proposed framework and AST4AV \cite{ast4av2018} through analyzing the velocity of the ego vehicle and the lane index in the scenario of a collision.

\textbf{Vehicle speed.}
\begin{figure}[htbp]
    \centering
    \begin{tikzpicture}
    \begin{axis}[
    width=0.95\linewidth,
    ybar,
    enlargelimits=0.2,
    legend style={
    legend columns=-1},
    ylabel={Percentage (\%)},
    xlabel={System Under Test's speed (m/s)},
    ylabel style={font=\large},
    xtick=data,
    xticklabels={0-5,5-10,10-15,15-20,20-25},
    xticklabel style={
      rotate=45
    },
    ]
    \addplot[ybar,pattern=dots]
    coordinates{
       (1,6.524184476940382)
       (2,19.23509561304837)
       (3,55.455568053993254)
       (4,16.08548931383577)
       (5,2.699662542182227)
    };
    \addplot[ybar, darkgray!20!black,fill=darkgray!80!white]
    coordinates {
      (1,0.28763183125599234)
      (2,3.231064237775648)
      (3,25.99232981783317)
      (4,33.969319271332694)
      (5,36.519654841802493)
    };
    \legend{AST4AV \cite{ast4av2018},\textbf{Our study}}
    \end{axis}
    \end{tikzpicture}
    \caption{Comparison of our proposed framework with AST4AV \cite{ast4av2018} in percentage of crashed scenario on velocity range.}
    \label{fig:velocity_analysis}
\end{figure}
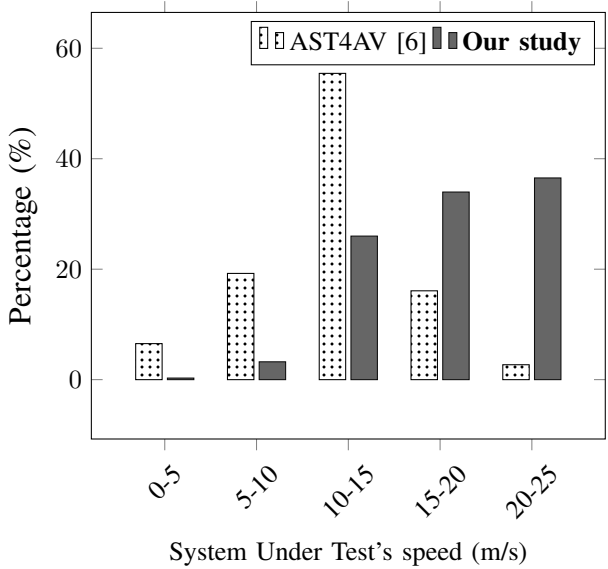
We first analyze the velocity of the SUT immediately preceding the crash scenario. In our observations, we grouped speed into five ranges: 0-5, 5-10, 10-15, 15-20, and 20-25 (m/s). The column chart for percentage of collision by velocity range is shown in Figure~\ref{fig:velocity_analysis}. The figure shows that crashed scenarios found by AST4AV are mostly distributed within the speed range of 10-15 (m/s). The percentage of velocity in the ranges 5-10 and 15-20 is approximately the same, while the percentage of velocity in the range 20-25 is smallest. It is because AST4AV \cite{ast4av2018} controls the action of surrounding vehicles, allowing the SUT to avoid running at high speeds on the multi-lane road. In our proposed framework, the crashed scenario occurred mostly in the highest velocity range, \ie, 20-25 (m/s), and the lower velocity range follows the percentage ranking. Since our framework is capable of capturing the crashes with higher speed, it is more reliable in real-world applications. 

\textbf{Lane change patterns.}
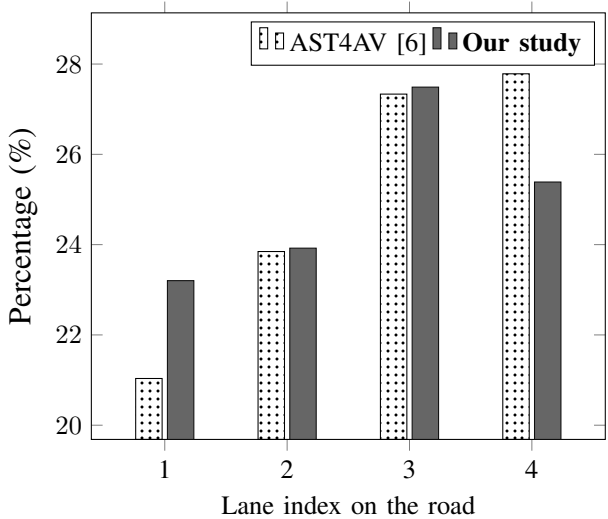
\begin{figure}[htbp]
    \centering
    \begin{tikzpicture}
    \begin{axis}[
    width=0.95\linewidth,
    ybar,
    enlargelimits=0.2,
    legend style={
    legend columns=-1},
    ylabel={Percentage (\%)},
    xlabel={Lane index on the road},
    ylabel style={font=\large},
    xtick=data,
    xticklabels={1,2,3,4},
    ]
    \addplot[ybar,pattern=dots]
    coordinates{
       (1,21.034870641169853)
       (2,23.847019122609673)
       (3,27.33408323959505)
       (4,27.78402699662542)
    };
    \addplot[ybar, darkgray!20!black,fill=darkgray!80!white]
    coordinates {
      (1,23.202301054650047)
      (2,23.921860019175455)
      (3,27.488974113135185)
      (4,25.38686481303931)
    };
    \legend{AST4AV \cite{ast4av2018},\textbf{Our study}}
    \end{axis}
    \end{tikzpicture}
    \caption{Comparison of our proposed framework with AST4AV \cite{ast4av2018} in percentage of crashed scenario on lane index.}
    \label{fig:lane_analysis}
\end{figure}
By analyzing the lane change index, we can discover the behaviors of lane changes from one lane to another. This data is useful for comprehending the lateral movements of autonomous vehicles during on-road testing. To that end, we quantify the proportion of crashed scenarios in each lane in both frameworks. The percentages of collisions by different lane are shown in Figure ~\ref{fig:lane_analysis}. For the second and third lanes, our framework achieves the same collision percentage as AST4AV \cite{ast4av2018}. In AST4AV, the collision percentage in the fourth lane, however, is higher than that in the 1st lane, causing an imbalance in the remaining two lanes. In general, the distribution of collision situations across each lane index in our framework is more evenly distributed, indicating that failures occurred at a variety of lane indexes.

\textbf{Impact of the reward function.}
\begin{table}[htbp]
\caption{Percentage of collision, between surrounding vehicles (Non-Ego) and only ego vehicle with surrounding vehicles (Ego) from different components in the reward function of our proposed framework.}
\label{table:components_comparison}
\begin{center}
\begin{tabular}{|l|r|r|r|r|r|r|} \hline
\multirow{3}{*}{\textbf{\textbf{Our study}}} & \multicolumn{6}{c|}{\textbf{Collision}}  \\ \cline{2-7} 
 & \multicolumn{2}{c|}{\textbf{Non-Ego}} & \multicolumn{2}{c|}{\textbf{Ego}} & \multicolumn{2}{c|}{\textbf{Total}}  \\ 
\cline{2-7} 
& \# & \%  & \# & \%  & \# & \%  
\\ \hline\hline 
$\lambda=0$ & 11 & 1.44\%            & 755 & 98.56\%        & 766 & 100\%   \\ \hline
$\lambda=1$   & 238 & 26.8\%           & 650 &  73.2\%   & 888 & 100\%   \\ \hline
$\lambda=0.8$        & 223 & 22.01\% & 864 &        77.99\%        & 1,087 & 100\% \\ \hline
\end{tabular}
\end{center}

\end{table}
We conduct an in-depth analysis to demonstrate the impact of each component on the reward function of our framework.
In this analysis, we observe a collision between a surrounding vehicle and surrounding vehicle. Our intuition is that if more collisions occur between surrounding vehicles, the surrounding vehicles will stop quickly in simulation, reducing the number of surrounding vehicles moving on the road and having a negative impact on finding failure scenario of SUT.
All collision scenarios can be divided into two types: (i) only one collision of the ego vehicle with the surrounding vehicle (Ego), and (ii) both collisions of the ego vehicle with the surrounding vehicle and collisions of the surrounding vehicle with the surrounding vehicle (Non-Ego).
Table \ref{table:components_comparison} presents the number of collisions and the percentage of each collision type using different weightings in the reward functions of the proposed framework. The results show that when only the safety of surrounding vehicles is considered (\ie, $\lambda=0$), the ratio of Non-Ego collision types is the smallest. It also yields the smallest number of collisions with the ego vehicle (\ie, 766). When only collision of the ego vehicle is considered (\ie, $\lambda=1$), the framework does not pay attention to the surrounding vehicles, and the number of Non-Ego type collisions is highest. As the number of collision types increases, the surrounding vehicle may stop earlier than the length of the episode, resulting in a smaller number of crashes of the ego vehicle at the end. The combination of both these considerations in reward function of our proposed framework enables us to find a larger amount of collisions compared to the existing methods in the literature (\ie, 1,087 versus 766 and 888). This demonstrates the advantages and flexibility of our framework in finding corner scenarios.


\begin{figure*}[tbp]
\begin{center}
\begin{tabular}{@{\hskip0pt}c@{\hskip5pt}c@{\hskip5pt}c@{\hskip5pt}c@{\hskip0pt}}

 AST4AV\cite{ast4av2018} & \textbf{Our study}, $\lambda=1$ & \textbf{Our study}, $\lambda=0$ & \textbf{Our study}, $\lambda=0.8$\\

\includegraphics[width=.2\linewidth,valign=m]{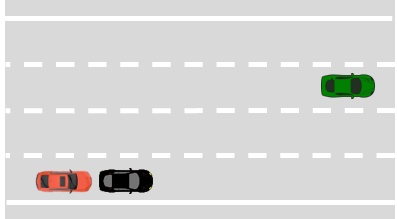} 
& \includegraphics[width=.2\linewidth,valign=m]{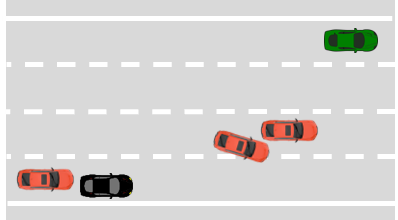} 
& \includegraphics[width=.2\linewidth,valign=m]{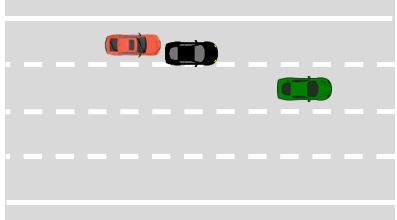} 
& \includegraphics[width=.2\linewidth,valign=m]{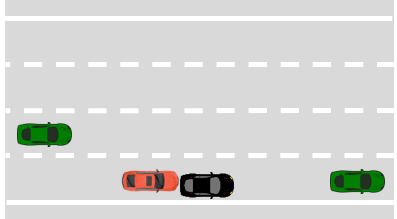}\\
\footnotesize{RE-a} & \footnotesize{RE-b} & \footnotesize{RE-c} & \footnotesize{RE-d}\\

\includegraphics[width=.2\linewidth,valign=m]{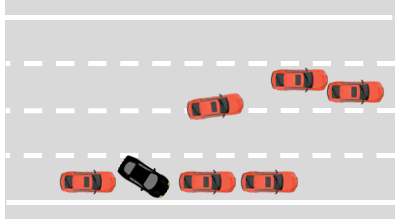} 
& \includegraphics[width=.2\linewidth,valign=m]{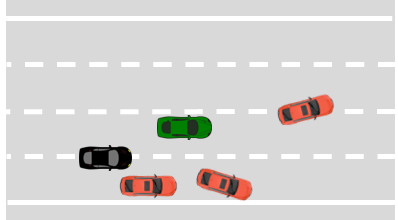} 
& \includegraphics[width=.2\linewidth,valign=m]{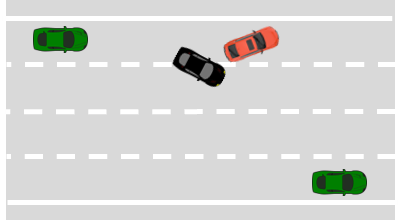} 
& \includegraphics[width=.2\linewidth,valign=m]{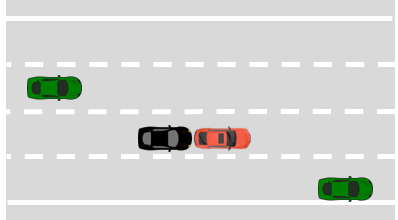}\\
\footnotesize{FE-a} & \footnotesize{FE-b} & \footnotesize{FE-c} & \footnotesize{FE-d}\\

\includegraphics[width=.2\linewidth,valign=m]{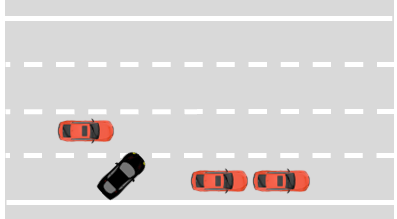} 
& \includegraphics[width=.2\linewidth,valign=m]{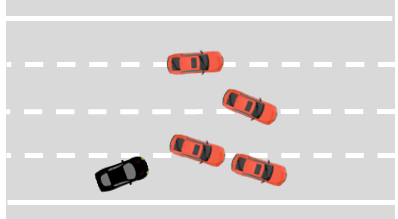} 
& \includegraphics[width=.2\linewidth,valign=m]{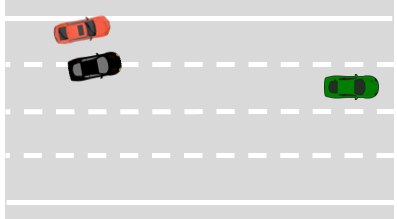} 
& \includegraphics[width=.2\linewidth,valign=m]{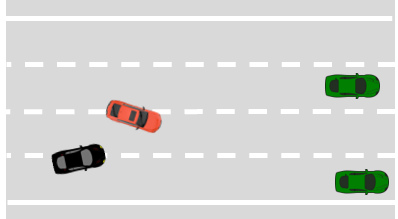}\\
 \footnotesize{CLL-a} & \footnotesize{CLL-b} & \footnotesize{CLL-c} & \footnotesize{CLL-d}\\

\includegraphics[width=.2\linewidth,valign=m]{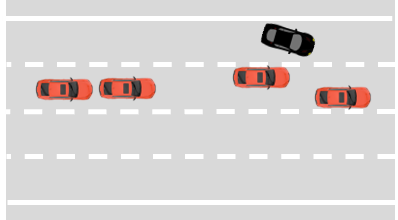} 
& \includegraphics[width=.2\linewidth,valign=m]{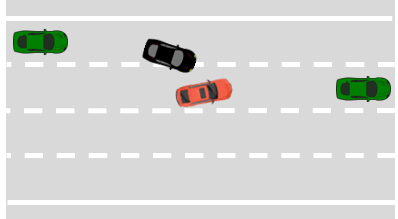} 
& \includegraphics[width=.2\linewidth,valign=m]{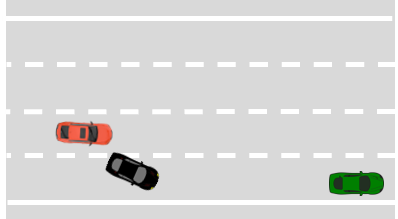} 
& \includegraphics[width=.2\linewidth,valign=m]{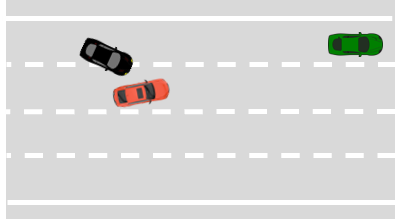}\\
\footnotesize{CRL-a} & \footnotesize{CRL-b} & \footnotesize{CRL-c} & \footnotesize{CRL-d}\\

\includegraphics[width=.2\linewidth,valign=m]{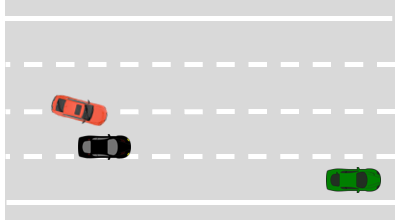} 
& \includegraphics[width=.2\linewidth,valign=m]{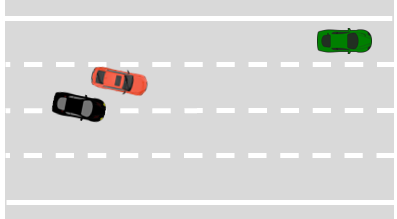} 
& \includegraphics[width=.2\linewidth,valign=m]{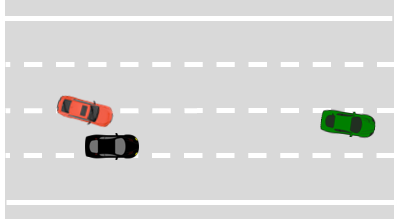} 
& \includegraphics[width=.2\linewidth,valign=m]{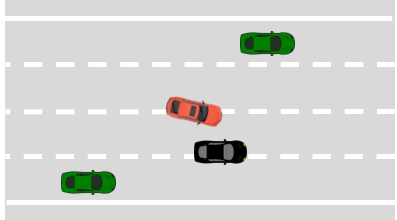}\\
\footnotesize{LVC-a} & \footnotesize{LVC-b} & \footnotesize{LVC-c} & \footnotesize{LVC-d}\\

\includegraphics[width=.2\linewidth,valign=m]{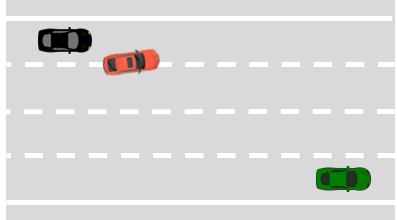} 
& \includegraphics[width=.2\linewidth,valign=m]{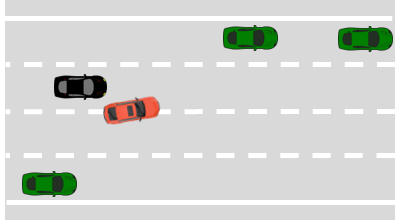} 
& \includegraphics[width=.2\linewidth,valign=m]{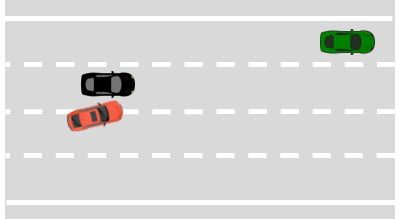} 
& \includegraphics[width=.2\linewidth,valign=m]{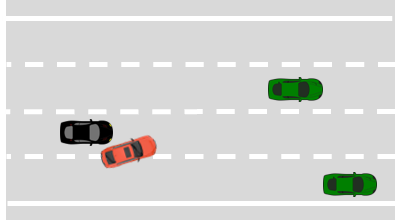}\\
\footnotesize{RVC-a} & \footnotesize{RVC-b} & \footnotesize{RVC-c} & \footnotesize{RVC-d}\\
\end{tabular}
\end{center}
\caption{
Examples of crashed scenarios sampled from various collision type for AST4AV \cite{ast4av2018} and our proposed framework. 
Vehicles are moving from left to right. \protect\tikz \protect\node [rectangle,draw, fill = black!70] at (2.5,-2) {};: Ego vehicle, \protect\tikz \protect\node [rectangle,draw, fill = red!70] at (2.5,-2) {};: Crashed vehicle(s), and \protect\tikz \protect\node [rectangle,draw, fill = green!70] at (2.5,-2) {};: Moving vehicle(s). {{\it RE}: collision occurs in the rare of ego vehicle; {\it FE}: collision occurs in the front of ego vehicle; {\it CLL}: collision occurs when ego vehicle attempt changing to the left lane; {\it CRL}: collision occurs when ego vehicle attempt changing to the right lane; {\it LVC}: collision occurs when a vehicle is coming from the left side of ego vehicle; {\it RVC}: collision occurs when a vehicle is coming from the right side of ego vehicle.}
}
\label{fig:qualitative_analysis}
\vspace{0.6cm}
\end{figure*}

\subsection{Discussions}\label{section:discussion}
We randomly sample examples of various collision types around the ego vehicle for the qualitative analysis. Figure~\ref{fig:qualitative_analysis} shows the crashed scenarios resulting from our proposed framework with variety of $\lambda$ values, and that of AST4AV \cite{ast4av2018}, respectively. Observe that the failure scenario generated by AST4AV and our study with $\lambda=1$ includes collisions of surrounding vehicles, aside from the collision between the ego vehicle and a surrounding vehicle. The collisions with other vehicles other than the ego vehicle may prevent the identification of further important failure scenarios. Reinforcing safe driving of the surrounding vehicles provides the opportunity for the algorithm to focus more on the collisions involving the ego vehicle. It can be seen in Figure~\ref{fig:qualitative_analysis} that our model with $\lambda = 0$ or $0.8$ generates scenarios with no collision among the surrounding vehicles and presents a more realistic driving conditions. The $\lambda=0.8$ value in our model produces the most realistic failure scenarios for all six observed crash types. In the RE type, the ego vehicle was slowing down to avoid colliding with the close leading vehicle in the same lane and at the same time was unable to change to the left lane due to another approaching vehicle resulting in the collision. In the FE type, the ego vehicle was not able to change lane (left or right) due to the close proximity of other vehicles in those lanes and thus involved in a collision when the leading vehicle suddenly stopped. In both CLL and CLR types, the ego vehicle changes lane to avoid the decelerated leading vehicle on that lane and is then involved in a collision with other vehicle who switches to the same lane simultaneously. In the LVC and RVC types, a vehicle other than the ego vehicle changes lane and collides with the ego vehicle on its lane. These scenarios demonstrate that our framework, that incorporates safety movement of surrounding vehicles generates the most efficient and realistic failure scenario where collision can happen in natural drivings.


\section{Conclusion}\label{section:conclusion}
In this paper, we proposed a novel AST framework to stress test the AV in a complex multi-lane road environment which includes a new reward function to encourage safe driving among other vehicles and support the identification of potential crashed scenarios. We developed a unified intelligent driving model (uIDM) that facilitates the movement of AV in both longitudinal and lateral directions which is much more realistic than models used in the existing literature. We calibrated the framework using data observed from California’s accident reports and then assess its performance against the existing IDM model to show the effectiveness of the proposed framework. Quantitative and qualitative analyses of our results demonstrated that our framework outperformed the existing state-of-the-art AST scheme in identifying corner cases with complex driving maneuvers. For future work, it is planned to extend our stress testing method to allow for the integration of end-to-end system and the consideration of more complex infrastructure and behaviours.

\section*{Acknowledgement}
This work was supported by the Australian Research Council (ARC) Discovery Grant DP190102134 and Industrial Transformation Research Hub IH180100010 (SPARC Hub).

\bibliographystyle{IEEEtran}

\bibliography{references.bib}

\ifpreprint
\else

\fi 
\end{document}